# Idle Vehicle Relocation Strategy through Deep Learning for Shared Autonomous Electric Vehicle System Optimization

Seongsin Kim[1]

Department of Mechanical Systems Engineering, Sookmyung Women's University,

Seoul 04310, South Korea

kss@sm.ac.kr

Ungki Lee[1]

Department of Mechanical Engineering, Korea Advanced Institute of Science and Technology,

Daejeon, 34141, South Korea

lwk920518@kaist.ac.kr

Ikjin Lee[*]

Department of Mechanical Engineering, Korea Advanced Institute of Science and Technology,

Daejeon, 34141, South Korea

ikjin.lee@kaist.ac.kr

Namwoo Kang[*]

Department of Mechanical Systems Engineering, Sookmyung Women's University,

Seoul 04310, South Korea

nwkang@sm.ac.kr

[1]Contributed equally to this work.

[*]Corresponding authors

**Abstract**

In optimization of a shared autonomous electric vehicle (SAEV) system, idle vehicle relocation strategies are important to reduce operation costs and customers' wait time. However, for an on-demand service, continuous optimization for idle vehicle relocation is computationally expensive, and thus, not effective. This study proposes a deep learning-based algorithm that can instantly predict the optimal solution to idle vehicle relocation problems under various traffic conditions. The proposed relocation process comprises three steps. First, a deep learning-based passenger demand prediction model using taxi big data is built. Second, idle vehicle relocation problems are solved based on predicted demands, and optimal solution data are collected. Finally, a deep learning model using the optimal solution data is built to estimate the optimal strategy without solving relocation. In addition, the proposed idle vehicle relocation model is validated by applying it to optimize the SAEV system. We present an optimal service system including the design of SAEV vehicles and charging stations. Further, we demonstrate that the proposed strategy can drastically reduce operation costs and wait times for on-demand services.

***Keywords:*** Idle vehicle relocation, deep learning, shared autonomous electric vehicle (SAEV), demand prediction, system optimization

# 1. Introduction

Shared autonomous electric vehicles (SAEVs) that combine car sharing services, autonomous driving technology, and electric vehicles (EVs) are expected to revolutionize transportation systems in the near future [1,2]. An SAEV autonomously goes to the location requested by a customer and rides that customer to a prescribed destination, thus providing a low-stress and safe transportation service [3,4], promoting transportation accessibility [5], and reducing mobility costs [6]. In addition, EVs help reduce fuel consumption and produce less environmental pollutants and greenhouse gas emissions [7–10]. Consequently, several studies related to operation and optimization of SAEV systems were conducted on the basis of SAEV simulations [1,11-15].

A primary issue regarding SAEV operation is the behavior or movement of idle vehicles. Most SAEV studies randomly selected moving strategies or implemented free-floating systems where the vehicles park at the current location or at a designated parking spot [15–19]. However, during practical SAEV operation, parking in the middle of the road is unrealistic. It is more efficient to move in advance to a location where passenger demands are expected than to move the vehicle around randomly. Further, customers' wait time can be reduced by predicting the passenger demands and preemptively moving idle vehicles to an appropriate destination, resulting in reduction of the required fleet size, the number of charging stations (CSs), and costs [20–27]. Previous research related to this study is reviewed in Section 2.

Previous studies have the following limitations. First, several of them proposed an idle vehicle relocation strategy based on optimization but there was a lack of research using deep learning. To embody on-demand service it is necessary to implement real-time decision-making based on big data. Therefore, deep learning is required to approximate a complex optimization process and reduce the computational cost. Second, several previous studies did predict passenger demand using deep learning; however, studies on SAEV system optimization using these predictive models are limited. SAEV system optimization can simulate the cost savings of idle vehicle relocation and determine the optimal vehicle size, vehicle specifications, and number and location of CSs.

This study proposes a deep learning-based idle vehicle relocation strategy and presents an SAEV vehicle and CS design framework that can minimize SAEV operation costs using the proposed strategy. The effectiveness of the proposed framework is verified by utilizing big data for taxi services in Seoul. The rest of this manuscript is structured as follows. In Section 3, the proposed framework is introduced. In Section 4, each step of the proposed framework is described in detail. The SAEV system optimization problems and models are discussed in Section 5. In Section 6, the simulation results are presented and discussed. Finally, Section 7 concludes the study and provides directions for future work.

# 2. Related Works

## 2.1. SAEV System Optimization

Several studies to optimize the SAEV fleet operation based on simulation were conducted. Chen et al. simulated the performance characteristics of SAEVs that provide service to travelers all over Austin, Texas [11]. They studied the extent to which the response time can be improved by reducing the charging time or increasing the number of vehicles. Kang et al. presented an SAEV system optimization framework integrating vehicle design, charging station locations, as well as fleet size and assignment [12]. This study compared the optimization results of an SAEV with those of a shared autonomous vehicle using internal combustion engine. Farhan et al. proposed an SAEV simulation framework that integrated optimization and discrete events and solved the routing optimization problem [13]. Zhao et al. proposed a station-based SAEV system model for simultaneous optimization of CS relocation and fleet size [14]. Loeb et al. applied a sizing strategy about a specific vehicle based on response time, empty vehicle-miles-travelled, and replacement rate [19]. They demonstrated that high-speed charging and long-range EVs are better options than shorter ranges and slow-charging vehicles. Lee et al. considered the uncertainties in SAEV systems and introduced reliability-based design optimization to derive an optimal SAEV system design that minimizes the total cost while satisfying the target reliability of the customer's wait time [15]. Melendez et al. proposed a cyber-physical system (CPS) comprising a large fleet of SAEVs and a set of charging hubs for optimal planning and real-time operations [28].

## 2.2. Idle Vehicle Relocation Strategy

Several studies were conducted on the positioning of idle vehicles, which have a considerable influence on the future demand and operation of SAEVs. Phithakkitnukoon et al. studied error-based learning through an inference engine based on a naïve Bayesian classifier and developed a prediction model for the number of empty taxis according to time and weather conditions [20]. Such a model exhibits a limitation as it does not reflect the real world well because it applies small-fleet samples. Yuan et al. proposed a parking-spot detection-based probability model algorithm that recommends passengers to access empty taxis using information about a high-income taxi driver's pickup action [21]. However, this model has the limitation that it does not integrate real-time traffic information. Li et al. transformed the digital tracking of Hangzhou 5350 taxis into a taxi pattern table and training/testing data set [22]; thus, they developed a taxi pattern determination algorithm for efficient passenger search through data mining algorithms. Zhang et al. proposed a dynamic bike relocation methodology to transform the model into an equivalent mixed-integer problem model and a heuristic algorithm for solving dynamic bicycle repositioning problem (BRP) [3]. Sayarshad et al. proposed a queueing-based formula about the idle vehicle relocation problem [24]. In this study, a Lagrange-decomposition heuristic algorithm that estimates the arrival process parameters for simulation using data from five days a week for six nodes was proposed and compared to a relaxed lower bound solution. Wen et al. used a model-free reinforcement learning approach that adopted a deep Q network for idle vehicles to resolve the imbalance between vehicle supply and passenger demand [25]. It was applied to an agent-based simulator and tested in a case study at London. Ma et al. proposed a ridesharing strategy using queueing theoretical vehicle dispatch and an idle vehicle relocation algorithm with integrated transit that picks up the passengers at home or at a transit station [26]. Pouls et al. proposed a prediction-based relocation algorithm using a mixed-integer programming model for idle vehicle relocation [27]. Real-time data sets from Hamburg, New York City (NYC), and Manhattan were used to demonstrate the suitability of this algorithm for real-time use.

## 2.3. Passenger Demand Prediction by Deep Learning

Deep learning-based studies that predict passenger demand for shared vehicles such as taxis achieved good performance. DeepST extracted spatiotemporal features by preprocessing the taxi demand data into an image formed as a grid array and inputting such image in a convolution layer [23]. The prediction performance was improved by using external information such as weather data; however, identifying similar patterns that repeat at specific cycles is difficult. DMVST-Net extracted the spatiotemporal and semantic contexts by converting $20\times20$ images into small $7\times7$ images and applying a local convolutional neural network (CNN), long short-term memory (LSTM), and graph embedding that use taxi demand data in Guangzhou City [29]. They extracted the semantic context through large-scale information network embedding (LINE), which is a graph embedding method. STDN used data in a particular time slot together with different data extracted in the past, beyond the past 4 hours of NYC's taxi data [30]. However, as the time range of input data widened, the gradient vanishing phenomenon intensified and training became difficult. The problem was solved by training the slot that has to concentrate with the periodically shifted attention mechanism, which is a temporal attention method. The TGNet model used the pickup and drop-off data of NYC taxi and the hidden pattern of time data to carry out training using only temporal guided embedding [31]. Thus, representing time, day of the week, holiday or not, and the day before holiday or not, they could train complex spatiotemporal patterns with a simple architecture. Xu et al. predicted the station demand and shared bicycle demand in real time using an LSTM neural network [32]. They predicted the shared bicycle demand in Nanjing, China, and the passenger demand and mobility patterns of bicycle sharing without station. Chen et al. extracted the spatial-pattern distribution using a graph convolutional network (GCN), and the temporal feature of the traffic flow using LSTM; they performed predictions about road sections [33].

# 3. Proposed Framework

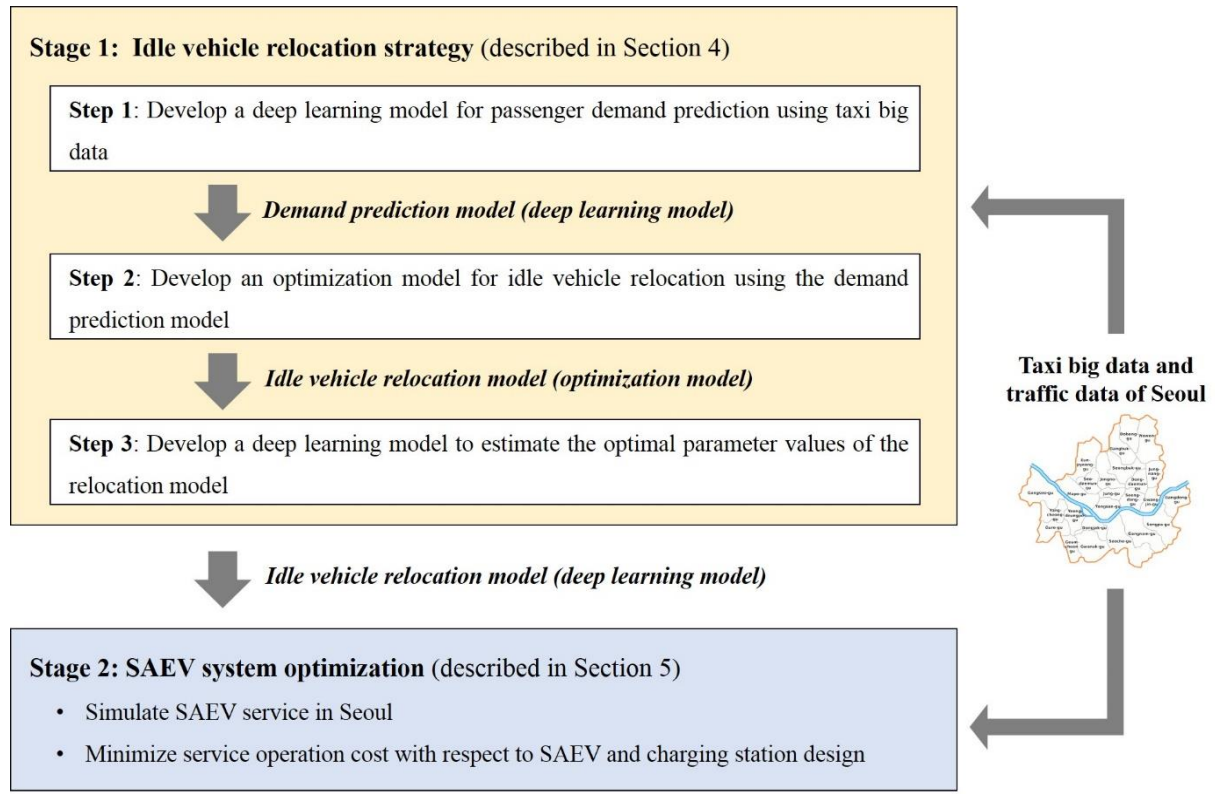

**Fig. 1.** Deep learning-based idle vehicle relocation strategy and SAEV system optimization.

The proposed framework entails two stages, as shown in Fig. 1. The data used for model development were the Seoul City's taxi-call big data provided by the Seoul Metropolitan Government (SEO-Taxi 2), collected over 15 months—from August, 2015, to October, 2016 [34].

Stage 1 proposes a deep learning-based idle vehicle relocation strategy. The deep learning model that trained the idle vehicle relocation solution was optimized for various traffic conditions in the past with end-to-end learning. Thus, it can decide the destination where the idle vehicle has to be placed according to given traffic conditions in the future. The proposed strategy takes place through Step 3 of the sequential model development.

Step 1 builds up a graph neural network (GNN)-based deep learning model that can predict the location and frequency of the passenger demand that will be known 30 minutes later based on the passenger demand location and frequency data; these data include information obtained from the previous 4 hours. In this study, we used the TGNet architecture and improved the performance by adding weather data as supplementary data [35].

Step 2 proposes an optimization model that decides the idle vehicle relocation based on passenger demand predicted through the demand prediction developed in Step 1. The model input comprises location and frequency of passenger demands, status of vehicles, and CS congestion, and the output is expressed in terms of the probability distribution of the idle vehicle destination.

Step 3 proposes a deep learning model that can decide the parameter values of the relocation model developed in Step 2 without going through optimization. Finding a solution by optimization based on various changing traffic conditions from big data is not suitable for on-demand service. Therefore, an approximation model that can propose an immediate solution as an alternative to the optimization model, which has high computational cost, is required. This step produces parameters $P_2$ and $P_3$, which are optimized according to the past traffic data used in Step 1, builds a data set, and conducts training with deep learning. The input data are transformed into 2D to represent the location information and are used for training a CNN-based model, which immediately calculates the values of $P_2$ and $P_3$.

In Stage 2, the proposed idle vehicle relocation strategy is applied to the SAEV system optimization problem. As a result, a service system design is proposed that can reduce the SAEV operation cost while satisfying the customer's wait time constraint. Using the deep learning prediction model and the idle vehicle relocation model developed in the previous stage, we simulated Seoul City's SAEV operation and determined the SAEV fleet size, battery design, location and number of CSs, and number of chargers in each station that can reduce the operation cost.

## 4. Idle Vehicle Relocation Strategy

### 4.1. Step 1: Passenger Demand Prediction by Deep Learning

The data used in the model for passenger demand prediction are the Seoul City's taxi call big data (08/01/2015~10/31/2016, 458 days). These data went through preprocessing, as shown in Fig. 2. We transformed the Seoul City's public taxi data set into a grid cell expression using the number of features in each grid (e.g., interest points and road segments) and the total length of the input features (e.g., road segments). To simplify the location information of passenger demand, we divided Seoul City's area into 2500 grids of 50×50; the area of each grid's cell was approximately 700×700 m$^2$. We mapped the passenger demand that took place in the corresponding grid cell by counting the requests. The value increases with the demand in the corresponding grid cell, exhibiting a brighter color as the value of the grid cell increases. Each cell represents the taxi-call GPS location and frequency within 30-min units. Each cell's passenger demand within such 30-min periods was added according to the cell locations. To predict the passenger demand that will be known 30 minutes later in the current standard based on a 4-h data period, the images corresponding to 30 minutes were transformed into a 2D image of 8 channels in total (4 h/30 min = 8). The resulting 8 overlapping images became the input data, whereas the 2D 50×50 image of one channel became the output data. Weather data were similarly transformed into 8-channel 2D 50×50 images. The data of the first 12 months were used for training, whereas the remaining 3 months became the test data.

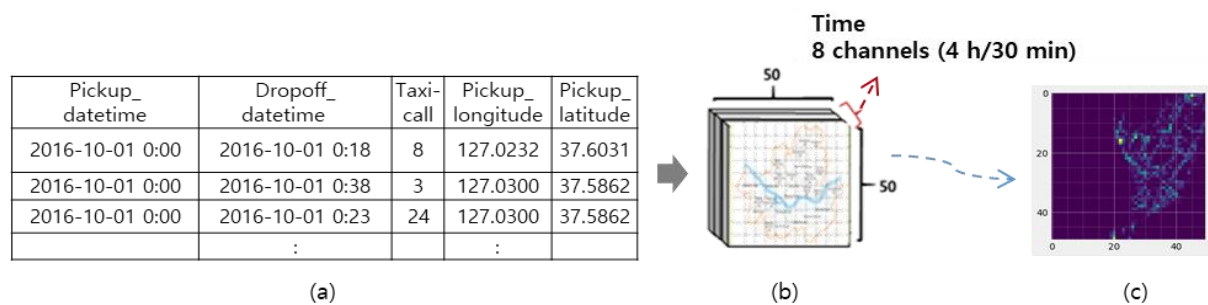

(a) Preprocessed public data set of Seoul city. Time and the road location where the taxi call occurred, the frequency of taxi calls and the passenger drop-off location.
(b) Map of Seoul is modeled by 700×700 m$^2$ grids. Frequency of taxi calls every 30 minutes is displayed as a grid, 50×50, combined channels.
(c) An example that showed the passenger demand in the grid cell into an image.

**Fig. 2.** Data preprocessing steps with Seoul taxi-call data.

For passenger demand prediction, we used TGNet, a deep learning model for prediction of taxi demand [35]. Additionally, trained the weather information of Seoul city by embedding. The deep learning architecture for training is shown in Fig. 3. In this study, the learning rate was set to 0.01, the decay rate was set to 0.01, the batch size was set to 128, the Adam optimizer was used as the optimizer, and early stopping was used to avoid overfitting. The training took approximately 8 h with four GPUs (GTX 1080) operating in parallel.

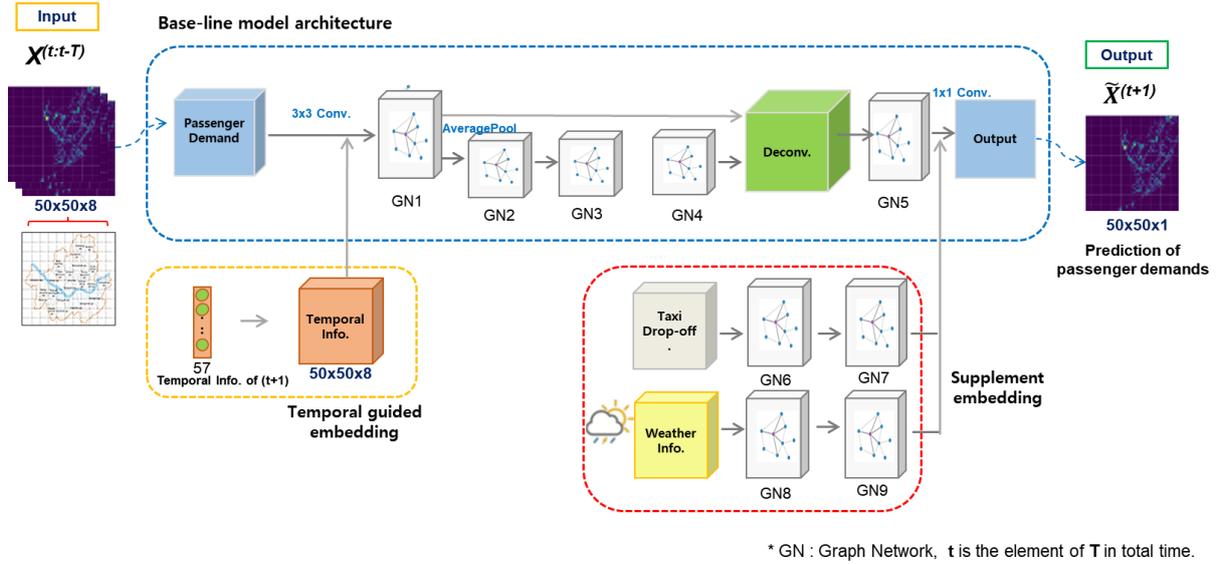

**Fig. 3.** TGNet model architecture (Base-line + **External Data**).

## 4.2. Step 2: Idle Vehicle Relocation by Optimization

The proposed idle vehicle relocation algorithm starts with selecting the destination of idle-state SAEVs by predicting passenger demands rather than applying random-motion strategies or free floating. The proposed algorithm is based on the passenger demand prediction model presented in Section 4.1. The flowchart of the algorithm is shown in Fig. 4. The passenger demand prediction model provides locations where the passenger demand is predicted to occur and the probability of occurrence ($P_1$). Thus, the candidates for potential destination can be specified. In addition to $P_1$, the algorithm considers the probability of selection according to the distance ($P_2$) and vehicle excess ($P_3$). $P_1$, $P_2$, and $P_3$ are determined by the predicted frequency of passenger demands, distance from the current location of the vehicle to the candidate, and number of vehicles already assigned that exceeds the predicted passenger demand, respectively. Then, $P_1 \times P_2 \times P_3$ is calculated for all candidates, and the candidate with the highest value is selected as the destination.

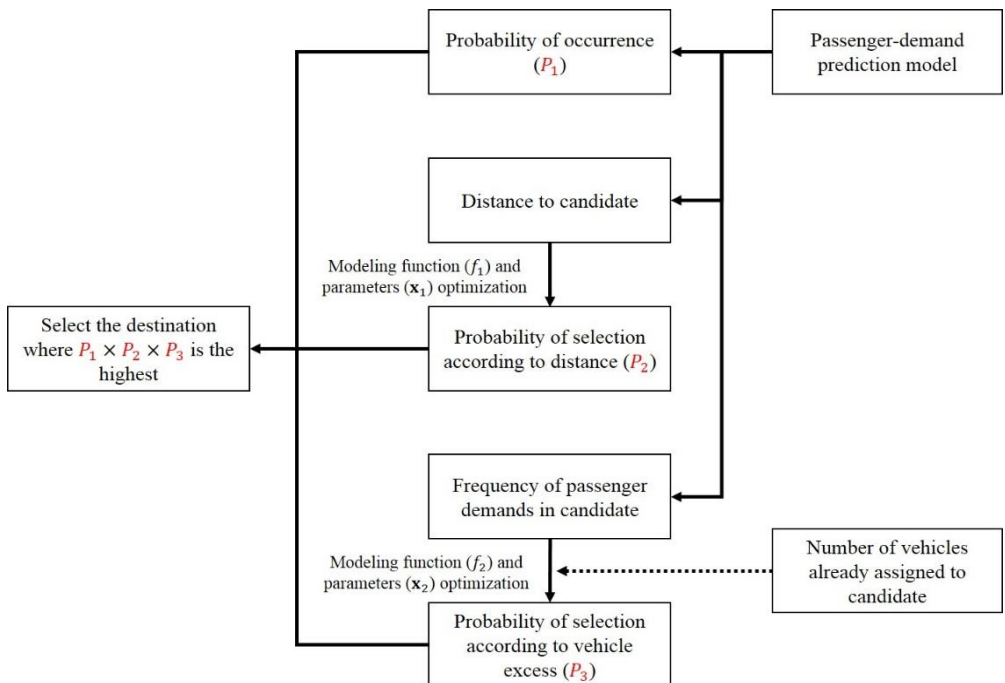

**Fig. 4.** Flowchart of the destination selection algorithm.

Modeling functions $f_1$ and $f_2$ are defined to control the probability of selection according to $P_2$ and $P_3$. Modeling functions work such that the candidate with closer distance and less vehicle excess has a high probability of selection. Distance and vehicle excess serve as variables of $f_1$ and $f_2$, respectively. Linear, concave, and exponential-Gauss functions summarized in Table 1 are used as candidates for modeling functions. The characteristics of the function vary depending on the parameters of each function—$\mathbf{x}_1$ is the vector of parameters for $f_1$ and $\mathbf{x}_2$ is the vector of parameters for $f_2$. The linear function has one parameter, whereas the concave and exponential-Gauss functions have two parameters. Within a given range of variables, the modeling functions take values between 0 and 1. Parameter bounds are set to express the diversity of a given modeling function sufficiently. For each modeling function, the variable $O_{ij}$ represents the distance from the current location $i$ to the destination $j$ for $f_1$, and the vehicle excess at the destination $j$ for $f_2$. The value of $f_2$ is 1 if no vehicle excess occurs. After the modeling functions are selected, parameters are optimized to minimize the mean customer wait time. Overall, the optimization problem can be formulated as follows:

$$\underset{\mathbf{x}_1, \mathbf{x}_2}{\arg\min} W(\mathbf{x}_1, \mathbf{x}_2) \tag{1}$$

where $W$ is the mean customer wait time obtained from SAEV simulation using the passenger demand prediction model and the destination selection algorithm.

**Table 1.** Modeling function types.

| Name | Function | Parameter bounds |
|---|---|---|
| Linear | $\max\{0, 1 - x_1 |O_{ij}|\}$ | $x_1$: [0,15] |
| Concave | $\max\{0, -x_1 |O_{ij}|^{x_2} + 1\}$ | $x_1$: [0,10], $x_2$: [1,10] |
| Exponential-Gauss | $e^{-x_1 |O_{ij}|^{x_2}}$ | $x_1$: [0,5], $x_2$: [0,5] |

$O_{ij}$ = Distance from $i$ to $j$ for $f_1$
    = $S_a^j - S_p^j$ for $f_2$

($S_a^j$ = number of assigned SAEV, $S_p^j$ = predicted frequency of passenger demands)

### 4.3. Step 3: Idle Vehicle Relocation by Deep Learning

As mentioned in Section 4.2, the optimal modeling functions and parameters of the destination selection algorithm can be obtained through optimization. However, changes in the conditions of an SAEV system, such as location and frequency of passenger demands, congestion of CSs, and status of vehicles, lead to changes in the optimal modeling functions and parameters. Therefore, this section presents a deep learning model that can operate under changing conditions of a given SAEV system by providing optimal modeling functions and parameters for given conditions in real time. As will be presented in Section 6.1.2, the exponential-Gauss function, which exhibits the best performance, is considered as the optimal modeling function type. Thus, only optimal modeling parameters of the exponential-Gauss function are of interest for deep learning models. The proposed deep learning process is shown in Fig. 5. After obtaining the optimal modeling parameters for given conditions of an SAEV system, the prediction model for modeling parameters can train the relationship between such conditions and optimal modeling parameters. In other words, the prediction model for modeling parameters can provide optimal modeling parameters when the conditions of an SAEV system at a certain moment are given.

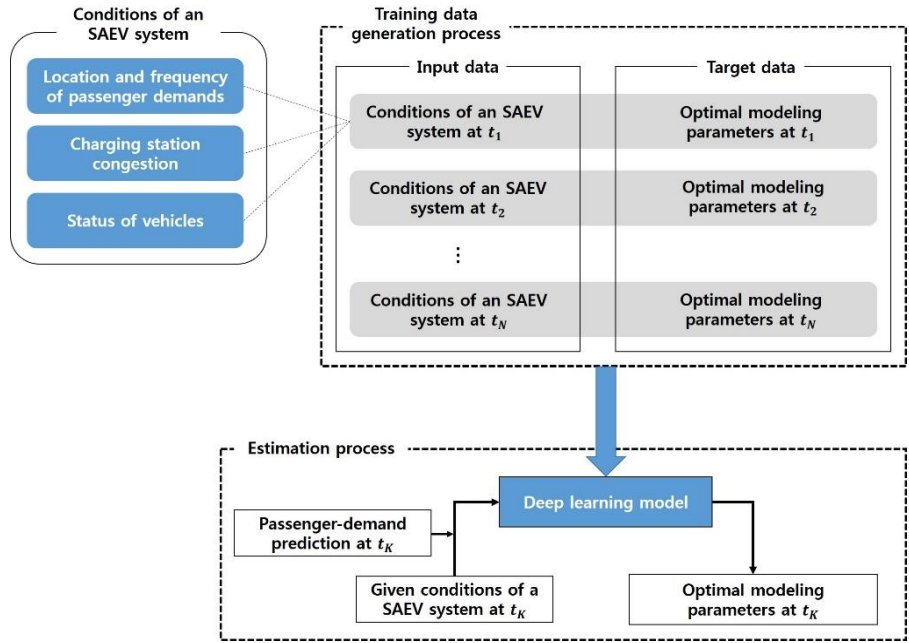

**Fig. 5.** Flowchart of the real-time prediction model for modeling parameters.

### 4.3.1. Training Data Generation

The training-data generation process produces input and target data needed for training the real-time prediction model for modeling parameters. Conditions of an SAEV system at a certain moment and corresponding optimal modeling parameters constitute the input and target data, respectively. In this study, real time was assumed to be a 30-min interval. Thus, SAEV simulations were conducted by dividing 24 hours per day into 30-min intervals to obtain training data. Fig. 6 shows an example of input and target data generation performed in a 30-min SAEV simulation at $t_i$. The conditions of an SAEV system comprise three factors used as the input data: (1) location and frequency of passenger demands for intervals of 30 minutes; (2) charging-station congestion indicating whether the chargers are occupied and when charging ends; and (3) status of vehicles, including current location, remaining battery capacity, and service status. The second and third factors are given at the beginning of the 30-min SAEV simulation. Then, optimal modeling parameters obtained through optimization from 30-min interval SAEV simulations become the target data of the interval. The consideration of three factors in training data allows the trained deep learning model to derive appropriate optimal modeling parameters in real time for given conditions. In this study, training data sets obtained from 30-min SAEV simulations for 363 days were used. Thus, a total of 17424 sets of input and target data were employed. To obtain the optimal modeling parameters from each simulation, a genetic algorithm was used for global search, and sequential quadratic programming was applied for local search.

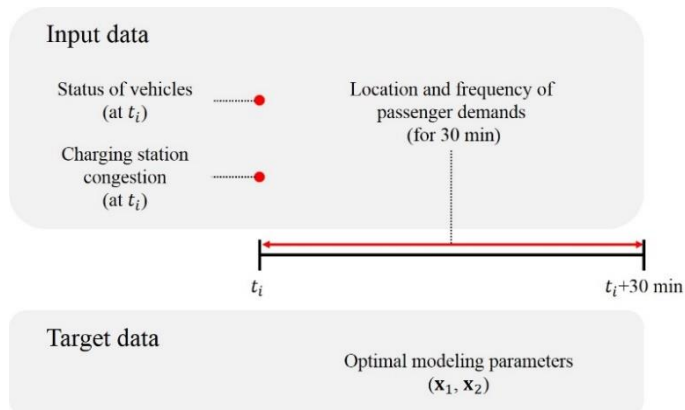

**Fig. 6.** Example of input and target data generation.

### 4.3.2. Training

The training data produced as described above were transformed into a 133×112 grid including all the location information to pass through the convolutional layer. This is one tenth of the area of Seoul city. In each channel and 4 channels of the charging station are combined, and 133×112×8 imaged data were used as input. Fig. 7 shows an example of preprocessed data: Fig. 7(a) shows the passenger demand's location and frequency for 30 minutes; Fig. 7(b) shows the CS congestion indicating whether the chargers are occupied and when charging ends; Fig. 7(c) shows the SAEV's current location; Fig. 7(d) shows the SAEV's battery capacity; and Fig. 7(e) shows the SAEV's service ending time. Each dot's color brightness is brighter as the corresponding number becomes greater.

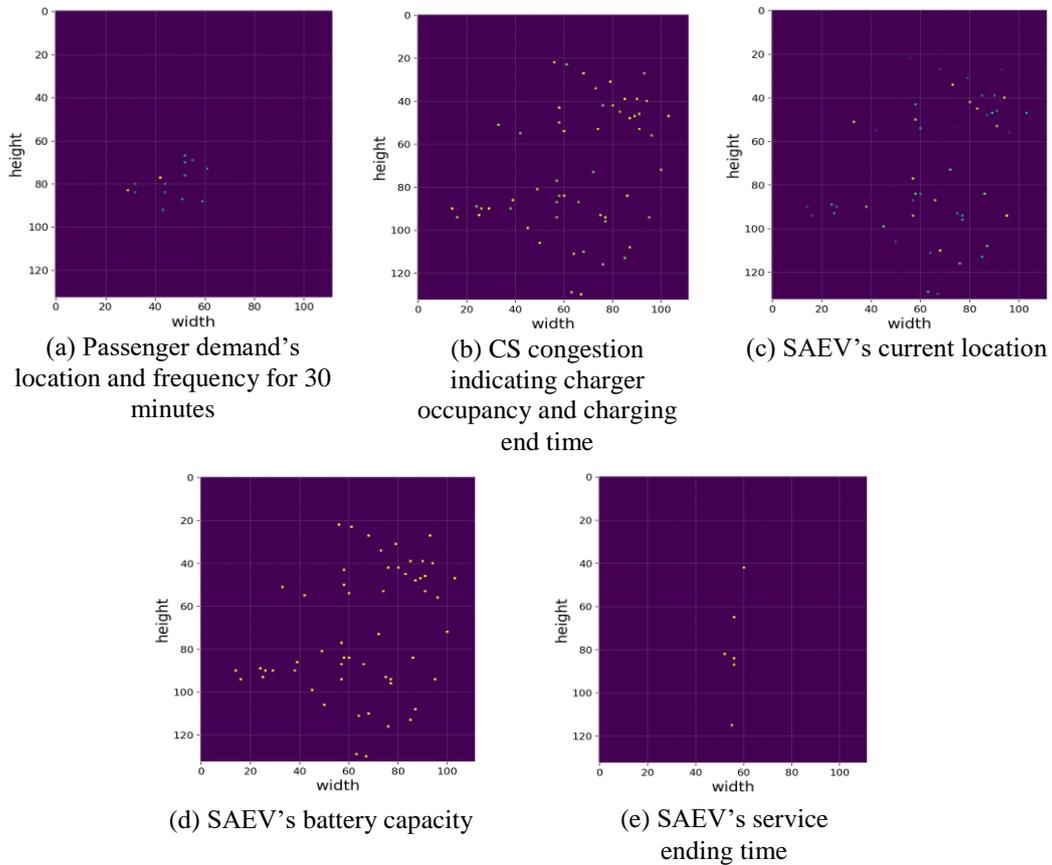

(a) Passenger demand's location and frequency for 30 minutes

(b) CS congestion indicating charger occupancy and charging end time

(c) SAEV's current location

(d) SAEV's battery capacity

(e) SAEV's service ending time

**Fig. 7.** Example of input data visualization as images.

Four parameter values in total were obtained at the output. Min-max scaling was performed on the input, and the logarithm was taken to increase the normality for target values and to derive meaningful results during regression analysis [36]. A total of 13839 data, which constitute 80% of 17424 data, were used as the training set, whereas 3484 data, that is, the remaining 20%, were used as the test set.

In this study, to determine the optimum architecture for mixed input features, we compared the performances of CNN, Unet [37], and GCN [38] architectures. We finally select the CNN-based architecture shown in Fig. 8. The proposed architecture is composed of 6 convolutional layers and batch normalization, 2 max-pooling layers, a ReLu activation function, and the Regressor part, which is composed of 1 fully connected layer and dropout. It was trained using the same Adam optimizer as for the passenger demand prediction model. The learning rate was set to 0.001; decay rate to 0.005; and batch size to 256; further, early stopping was applied. The training took approximately 1 hour with four GPUs (GTX 1080) operating in parallel.

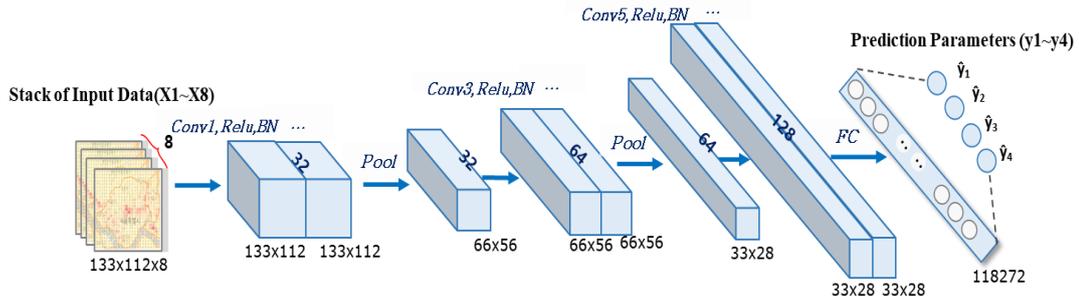

**Fig. 8.** Deep learning architecture for idle vehicle relocation.

## 5. SAEV System Optimization

In this study, the SAEV system framework presented by Lee et al., which follows a random-motion strategy, was modified to apply the proposed destination selection algorithm to Seoul's data [15]. The SAEV system framework presented in Fig. 9 consists of a fleet operation model, CS model, and SAEV design. It derives the customer's wait time for a given SAEV system design. The fleet operation model reflects road circumstances to perform shortest-time-path search and determines the optimal fleet assignment by considering the destination selection of vehicles not in service and the charging schedule. The CS model determines the optimal locations of CSs and the total number of chargers. Finally, the SAEV design model determines the charging time of the battery and EV performance metrics, such as driving range, top speed, acceleration, and miles per gallon equivalent (MPGe). In this study, fleet operation is assumed to be performed by a central operating system that manages passenger demands, destination selections, and status of EVs and CSs. Detailed fleet operation, CS network design, and SAEV design models are shown in Appendix.

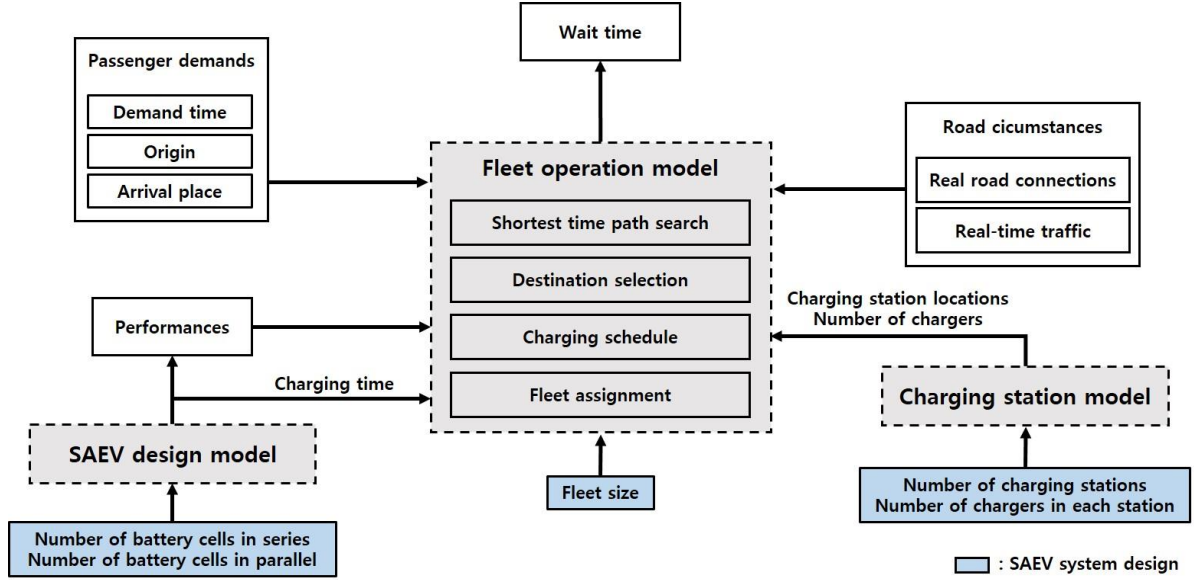

**Fig. 9.** SAEV system framework.

A seven-day simulation was conducted for the SAEV system optimization problem that minimizes the cost while satisfying the customer's wait time constraint, and the results are discussed in Section 6.2. In the seven-day simulation, we assumed that an SAEV occupies a part of the traditional transportation system and on average 942 customers use the SAEV service per day. The customer's passenger demand time, origin, and arrival place in the simulation were generated upon the passenger demand data of Seoul. The number of customers per day was appropriately determined considering the time required for simulation. The optimization problem for the design of the SAEV system is formulated as follows [15]:

$$\begin{aligned}
\text{find} \quad & \mathbf{X} = \left[ N_{\text{CS}}, N_{\text{charger}}, N_{\text{SAEV}}, \mathbf{X}_{batt}^{\text{T}} \right] \\
\min_{\mathbf{X}} \quad & \text{Cost}(\mathbf{X}) \\
\text{subject to} \quad & \mathbf{lb} \leq \mathbf{X} \leq \mathbf{ub} \\
& g_{\text{SAEV}}(W) \leq 0 \\
\text{where} \quad & \mathbf{X}_{batt} = [N_{\text{S}}, N_{\text{P}}]^{\text{T}} \\
& [P_{\text{range}}, T_{\text{charging}}] = f_{\text{SAEV}}(\mathbf{X}_{batt}) \\
& [\mathbf{L}_{\text{CS}}, C_{\text{CS}}] = f_{\text{CS}}(N_{\text{CS}}, N_{\text{charger}}) \\
& [W, C_{\text{SAEV}}] = f_{\text{operation}}(N_{\text{SAEV}}, \mathbf{L}_{\text{CS}}, N_{\text{charger}}, P_{\text{range}}, T_{\text{charging}}),
\end{aligned} \quad (2)$$

where the objective of the problem is to minimize the cost, consisting of CS and SAEV fleet costs; $\mathbf{X}$ is a decision variable vector; $N_{\text{CS}}$, $N_{\text{charger}}$, and $N_{\text{SAEV}}$ are the numbers of CSs and chargers per each CS, and the SAEV fleet size, respectively; $\mathbf{X}_{batt}$ indicates the battery design variable vector; $\mathbf{lb}$, $\mathbf{ub}$, and g are the lower and upper bounds, and the customer's wait time constraint, respectively; $W$ is the mean customer's wait time, whose constraint was set such that $W$ satisfies 5 minutes; $N_{\text{S}}$ and $N_{\text{P}}$ are the numbers of battery cells in series and parallel, respectively; $P_{\text{range}}$ and $T_{\text{charging}}$ are the driving range and charging time of an SAEV, respectively; $\mathbf{L}_{\text{CS}}$ denotes the CS location vector; $C_{\text{CS}}$ is the CS cost encompassing the installment and maintenance of CSs; $C_{\text{SAEV}}$ is the SAEV fleet cost; and $f_{\text{SAEV}}$, $f_{\text{CS}}$, and $f_{\text{operation}}$ indicate the SAEV design, CS, and fleet operation models, respectively. Before starting the optimization, the candidates for CS locations are pre-determined using a *p*-median model (see Appendix A2).

An SAEV incurs the following costs: lithium-ion battery, $236/kWh [39]; autonomous module, $10,000 [12]; 80-kW permanent magnet synchronous motor, $1,665 [40]; other costs for the vehicle, assumed to be $6,000. The installment and maintenance costs of each charger in CS are assumed to be $22,626 and $5,000, respectively [12,41].

## 6. Simulation Results

### 6.1. Idle Vehicle Relocation Strategy

#### 6.1.1. Deep learning-based Demand Prediction

Concerning performance metrics, the passenger demand prediction, root mean square error (RMSE), and mean absolute percentage error (MAPE) were used. To analyze the suitability of the TGNet model, we compared the performance with that of the NYC taxi data [31,42], which constitute the benchmarking data set. The results are summarized in Table 2. SEO-Taxi 1 is the Seoul-rite-hailing dataset of Kakao taxi used in TGNet, and SEO-Taxi 2 is the Seoul taxi vendor data set provided by Seoul City used after preprocessing. When weather data were used in SEO-Taxi 2 data, the performance in terms of both RMSE and MAPE was better than when weather data were not used.

**Table 2.** Comparison of forecasting results on NYC-taxi, SEO-taxi1, and SEO-taxi2 for the two network models.

| Method | NYC-Taxi | | SEO-Taxi 1 | | SEO-Taxi 2 | | SEO-Taxi 2 with External Data | |
|---|---|---|---|---|---|---|---|---|
| | RMSE | MAPE (%) | RMSE | MAPE (%) | RMSE | MAPE (%) | RMSE | MAPE (%) |
| TGNet | 22.75 | 14.83 | 25.35 | 35.72 | 3.90 | 33.20 | 3.67 | 31.60 |

#### 6.1.2. Idle Vehicle Relocation by Optimization

This section presents the optimization results of modeling functions and parameters constituting the destination selection algorithm explained in Section 4.2. The optimization results are compared in terms of the mean wait time of customers obtained from the seven-day simulation presented in Section 5. In the simulation, the SAEV system design was assumed to accommodate the given number of customers per day: the number of fleets was 63; number of charging stations and chargers in each station were 6 and 4, respectively; number of battery cells in series and parallel were 110 and 2, respectively; and gear ratio was 8.2. The range, MPGe, acceleration, and top speed obtained from the given SAEV design are 179.3 km, 135.9, 10.1 s, and 150.5 km/h, respectively. Real road connections and CS locations in Seoul are shown in Fig. 10 and the determination of CS locations is explained in Appendix A2. For the purpose of comparison, the mean wait time results obtained using a random-motion strategy, which was frequently applied in previous SAEV studies, were employed as benchmarks. The results are summarized in Table 3.

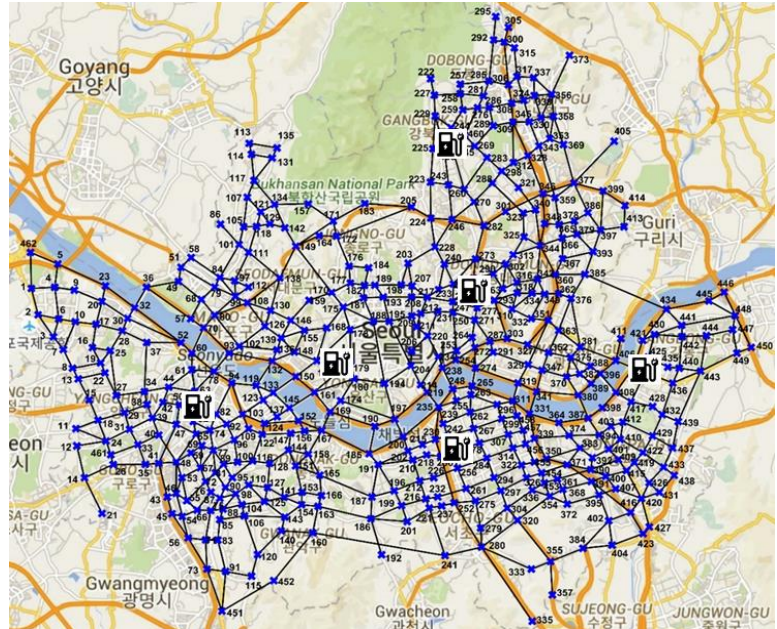

**Fig. 10.** Real road connections and CS locations in Seoul.

**Table 3.** Mean wait time results obtained using a random-motion strategy (minutes).

| Day 1 | Day 2 | Day 3 | Day 4 | Day 5 | Day 6 | Day 7 |
|---|---|---|---|---|---|---|
| 11.90 | 11.69 | 10.44 | 11.03 | 11.49 | 12.76 | 12.45 |

Table 4 summarizes the optimal parameters and mean wait time results according to the modeling function type. The given modeling function type was equally applied to $f_1$ and $f_2$, and the optimization problem expressed by Eq. (1) was solved for each day (24 hours) to derive optimal parameters. For optimization of this 24-h interval simulation, a genetic algorithm was used for global search, and sequential quadratic programming was used for local search. The computation time for one optimization was 4 hours on average using a standard desktop computer (Intel Xeon 8168 CPU @ 2.70 GHz with 192.0 GB of RAM).

**Table 4.** Optimization results depending on the modeling function type.

|  | Modeling function type | $f_1$ | | $f_2$ | | Mean wait time (minutes) |
|---|---|---|---|---|---|---|
|  |  | $x_1$ | $x_2$ | $x_3$ | $x_4$ |  |
| Day 1 | Linear | 0.1007 | - | 1.4064 | - | 3.24 |
|  | Concave | 0.0458 | 1.2487 | 0.3251 | 9.6157 | 3.22 |
|  | Exponential-Gauss | 0.1059 | 1.5015 | 3.0336 | 2.6666 | 3.16 |
| Day 2 | Linear | 0.0969 | - | 1.4650 | - | 3.01 |
|  | Concave | 0.0034 | 2.3840 | 6.2043 | 3.7056 | 3.02 |
|  | Exponential-Gauss | 0.0145 | 2.3180 | 3.6977 | 2.1203 | 2.95 |
| Day 3 | Linear | 0.1598 | - | 0.4219 | - | 2.58 |
|  | Concave | 0.0047 | 2.9645 | 0.1526 | 4.2942 | 2.57 |
|  | Exponential-Gauss | 0.3978 | 1.0491 | 0.9246 | 1.5538 | 2.54 |
| Day 4 | Linear | 0.1025 | - | 0.9983 | - | 2.81 |
|  | Concave | 0.0185 | 1.7005 | 5.6648 | 8.6433 | 2.80 |
|  | Exponential-Gauss | 0.0488 | 2.2345 | 1.7868 | 1.3224 | 2.76 |
| Day 5 | Linear | 0.0930 | - | 0.9634 | - | 2.88 |
|  | Concave | 0.0115 | 1.9886 | 0.8139 | 5.6652 | 2.87 |
|  | Exponential-Gauss | 0.1656 | 1.2461 | 2.1201 | 2.1727 | 2.85 |
| Day 6 | Linear | 0.0687 | - | 2.1172 | - | 4.27 |
|  | Concave | 0.0518 | 1.0867 | 7.9813 | 5.3493 | 4.16 |
|  | Exponential-Gauss | 0.0683 | 1.8004 | 4.9477 | 0.7728 | 4.16 |
| Day 7 | Linear | 0.0893 | - | 2.5576 | - | 3.62 |
|  | Concave | 0.0088 | 1.7094 | 2.0491 | 2.1507 | 3.59 |
|  | Exponential-Gauss | 0.0246 | 1.9897 | 0.5280 | 3.1947 | 3.55 |

For all modeling function types, the mean wait time obtained from the simulation using the destination selection algorithm with optimized parameters was considerably reduced with respect to the simulation using a random-motion strategy—the mean wait time was reduced by 72.73% (8.46 minutes), 72.93% (8.50 minutes), and 73.29% (8.54 minutes) on average when using the linear, concave, and exponential-Gauss functions, respectively. This shows that the destination selection algorithm properly selects the destination of the idle-state SAEV and efficiently performs the fleet operation. The mean wait time varied depending on the modeling function type—the mean wait time was 3.20 minutes, 3.18 minutes, and 3.14 minutes on average when using the linear, concave, and exponential-Gauss functions, respectively. This indicates that nonlinear functions describe the probability of selection better than a simple linear function. In addition, the exponential-Gauss function is more suitable than the concave function for modeling. This was confirmed in all seven-day simulations. As a result of performing optimization in a situation where the modeling function types of $f_1$ and $f_2$ can be different, the optimal modeling functions for both $f_1$ and $f_2$ were exponential-Gauss throughout the seven days. The results also demonstrate that the optimal parameters of the modeling function were different for each day, which means that the optimal parameters vary according to the conditions of a given SAEV system.

### 6.1.3. Idle Vehicle Relocation by Deep Learning

To evaluate the performance of the deep learning model, we used both RMSE and mean absolute error (MAE) as evaluation metrics. The RMSE and MAE for the test data were 1.25 and 0.93, respectively. To demonstrate the performance of the modeling parameter prediction model for real-time operation, the mean wait time results obtained by optimizing modeling parameters from 30-min interval simulations were used as benchmarks. The mean wait time and optimal modeling parameter results from seven-day simulations—each day's simulation consisting of 48 30-min simulations—are summarized in Table 5. As in Section 6.1.2, a genetic algorithm was used for global search, and sequential quadratic programming was applied for local search. The mean wait time obtained by optimizing modeling parameters from 30-min interval simulations was reduced by 21.07% (0.7 minutes) on average compared to that of the 24-h interval simulation. This indicates that it is important to provide optimal modeling parameters for each short interval simulation given that the optimal values change continuously depending on the given conditions of an SAEV system. The large standard deviation of optimal parameters also shows that the optimal parameters change significantly for every 30-min interval simulation. This stresses the need for an alternative method to optimization that promptly provides optimal modeling parameters, as it is difficult to complete optimization in a short time.

**Table 5.** Mean wait time and optimal parameter results from 30-min interval simulation (obtained through optimization).

|       | $f_1$ | | $f_2$ | | Mean wait time (minutes) |
|---|---|---|---|---|---|
|       | $x_1$ | $x_2$ | $x_3$ | $x_4$ |  |
| Day 1 | 1.2889 | 0.9365 | 2.6879 | 2.6654 | 2.44 |
|       | (1.1798)* | (0.9041) | (1.5383) | (1.4913) |  |
| Day 2 | 1.7521 | 1.1572 | 2.8835 | 2.7875 | 2.32 |
|       | (1.3620) | (1.1367) | (1.3148) | (1.4591) |  |
| Day 3 | 2.1739 | 1.6720 | 2.4483 | 2.0584 | 2.23 |
|       | (1.5006) | (1.5035) | (1.5440) | (1.2894) |  |
| Day 4 | 1.2818 | 1.2396 | 2.6553 | 2.4479 | 2.33 |
|       | (1.2234) | (1.2750) | (1.5672) | (1.4271) |  |
| Day 5 | 1.4679 | 1.2077 | 2.9140 | 2.3957 | 2.41 |
|       | (1.2103) | (1.1902) | (1.4210) | (1.3621) |  |
| Day 6 | 1.5970 | 0.9752 | 2.6074 | 2.2495 | 2.76 |
|       | (1.5075) | (1.1637) | (1.3903) | (1.4581) |  |
| Day 7 | 0.9561 | 0.9870 | 2.6718 | 2.4123 | 2.60 |
|       | (0.9634) | (1.1010) | (1.4152) | (1.5566) |  |

*Standard deviations are enclosed in parentheses

The modeling parameter prediction model for real-time operation provides optimal modeling parameters according to given conditions of an SAEV system every 30 minutes. The mean wait time and optimal modeling parameters are presented in Table 6. They were obtained from a seven-day simulation through 30-min intervals performed with optimal modeling parameters provided by the modeling parameter prediction model. When using such model instead of optimization, the mean wait time was increased by 5.40% (0.13 minutes) on average, but it was reduced by 77.97% (9.10 minutes) on average with respect to applying a random-motion strategy. This shows that the decrease in the customer's wait time was still significant when using the modeling parameter prediction model. This also indicates that the modeling parameter prediction model provides appropriate parameters according to the given SAEV system condition. The difference between the parameters provided by such model and the parameters obtained through optimization can be explained by the error generated in the predicted location and frequency of passenger demands, which are provided by the passenger demand prediction model and used as input data of the modeling parameter prediction model.

**Table 6.** Mean wait time and optimal parameter results from 30-min interval simulations (using the modeling parameter prediction model).

|       | $f_1$ | | $f_2$ | | Mean wait time (minutes) |
|---|---|---|---|---|---|
|       | $x_1$ | $x_2$ | $x_3$ | $x_4$ |  |
| Day 1 | 0.9964 | 0.7899 | 1.7234 | 2.0064 | 2.56 |
|       | (0.5149)* | (0.4564) | (0.8806) | (0.7399) |  |
| Day 2 | 1.2114 | 0.8852 | 1.9380 | 1.6121 | 2.48 |
|       | (0.6982) | (0.4664) | (0.9750) | (0.7479) |  |
| Day 3 | 1.2288 | 0.9018 | 1.7609 | 1.7782 | 2.30 |
|       | (0.6687) | (0.4693) | (0.7822) | (0.7563) |  |
| Day 4 | 1.1217 | 0.7963 | 1.9193 | 1.9347 | 2.41 |
|       | (0.6847) | (0.6451) | (0.9745) | (0.7885) |  |
| Day 5 | 0.9873 | 0.7337 | 2.0396 | 1.9675 | 2.55 |
|       | (0.5378) | (0.4624) | (0.9936) | (0.9201) |  |
| Day 6 | 0.9960 | 0.8238 | 2.1468 | 2.1951 | 2.97 |
|       | (0.5110) | (0.4729) | (0.9536) | (0.9066) |  |
| Day 7 | 0.9964 | 0.7899 | 1.7234 | 2.0064 | 2.76 |
|       | (0.5149) | (0.4564) | (0.8806) | (0.7399) |  |

*Standard deviations are enclosed in parentheses

## 6.2. SAEV System Optimization

Table 7 summarizes optimal designs and outcomes obtained from Eq. (2) where the destination selection algorithm using the modeling parameter prediction model for real-time operation is applied. The results are compared with those obtained using a random-motion strategy. When using the destination selection algorithm, the cost was reduced by 38.2% with respect to applying the random-motion strategy, demonstrating that the destination selection algorithm and modeling parameter prediction model for real-time operation work well. Owing to the modeling parameter prediction model, which provides optimal modeling parameters immediately according to the given conditions of an SAEV system, the destination selection algorithm can be successfully applied to the optimization problem. Through the modeling parameter prediction model, the optimal modeling parameters required in the destination selection algorithm can be provided in real time. Thus, the destination of the idle-state SAEV is properly selected and the efficiency of the fleet operation can be increased. This results in reduction in the fleet size and the number of charging stations required to satisfy the wait time constraint.

**Table 7.** Optimal designs and outcomes of the SAEV system optimization problem with wait time constraint.

| | | Random-motion strategy | Destination selection algorithm |
|---|---|---|---|
| Decision variables | Fleet size | 75 | 50 |
| | Number of CSs | 9 | 6 |
| | Location of CSs | 64,173,193,203,251,268, 379,424,459 | 64,173,251,268,424,459 |
| | Number of chargers | 3 | 2 |
| | Number of battery cells in series | 101 | 139 |
| | Number of battery cells in parallel | 1 | 1 |
| Vehicle spec. | Acceleration (0–100) | 10.3 s | 8.5 s |
| | Range | 84.3 km | 115.3 km |
| | Top speed | 148.2 km/h | 158.2 km/h |
| | MPGe | 140.0 | 138.5 |
| | Battery capacity | 12.7 kWh | 17.5 kWh |
| Cost | Total cost | $2,309,134 | $1,427,066 |
| | Fleet cost | $1,549,731 | $1,089,554 |
| | CS installment cost | $610,903 | $271,512 |
| | CS maintenance cost | $148,500 | $66,000 |
| Fleet operation | Charging time | 13.5 minutes | 18.6 minutes |
| Customer's wait time | Mean | 4.99 minutes | 4.94 minutes |

## 7. Conclusion

This study proposes a deep learning-based idle vehicle relocation strategy that decides the location of the idle SAEV in real time. The proposed idle vehicle relocation strategy is achieved through 3 steps. In Step 1, a deep learning-based prediction model is built to forecast the location and frequency of passenger demand. Interestingly, weather information is added to TGNet as supplemental data to improve the performance. In Step 2, a relocation optimization algorithm decides the idle vehicle destinations based on passenger demand prediction. Three probability distributions are defined for relocation strategy and for setting optimal modeling parameter values for training. In Step 3, a deep learning model estimates the optimal modeling parameters without optimization process. The proposed idle vehicle relocation strategy was also applied to the SAEV system optimization problem, and the performance of the modeling parameter prediction model for real-time operation was investigated. The proposed idle vehicle relocation algorithm reduced the cost by 38.2% with respect to applying a random-motion strategy. The relocation algorithm makes efficient automobile operation CS possible and reduces the number of charging stations and automobile size needed to fulfill the wait time limit. These results confirm that the proposed algorithm operates well for on-demand service.

Concerning future work, we plan to focus on improving the accuracy of modeling parameter prediction by taking into account various conditions, such as CS congestion and vehicle condition, and location and frequency of travel-request passenger demand, which will be used as training data for modeling parameter prediction models.

**Declaration of Competing Interests**

The authors declare that they have no known competing financial interests or personal relationships that could have influenced the work reported in this study.

**Acknowledgments**

This study was supported by National Research Foundation of Korea (NRF) grants funded by the Korean government [grant numbers 2017R1C1B2005266 and 2018R1A5A7025409] and Energy Cloud R&D Program through the National Research Foundation of Korea (NRF) funded by the Ministry of Science, ICT (No. 2019M3F2A 1072468). This study was also supported by HPC Support Project of the Ministry of Science and ICT and NIPA,

# Appendix

## A1. Fleet Operation Model

The fleet operation model determines the optimal fleet assignment according to passenger demands and provides customer's wait time as the output. In the fleet operation of an SAEV, the fleet can have three states: (1) idle, in which the vehicle is not in service and moves to the selected destination where passenger demands are expected to occur; (2) in-service, in which the vehicle is assigned to a customer and is occupied; and (3) charging, in which the battery is charged when the state-of-charge (SOC) of the battery falls below a certain level. The flowchart of the fleet operation model for 24 hours is presented in Fig. A.1. The initial vehicle information for each day, including the vehicle location, the SOC of the battery, and the time when the vehicle service ends and its state becomes idle, is set randomly. In the idle state, the fleet moves to the destination according to the proposed destination selection algorithm. The fleet can be considered as a candidate vehicle to be assigned to the customer when a passenger demand occurs during the route. The central operating system continuously monitors the SOC of the battery during the idle state. It instructs the vehicle to go to a CS when the SOC of the battery reaches a certain level. If empty chargers exist, the nearest charger is assigned, whereas the charger with minimum wait time is assigned in case all chargers in the CSs are occupied. The central operating system manages the charging schedule by monitoring the start and end times of charging for each charger. When a passenger demand occurs, a vehicle with feasible SOC to go to the nearest CS after service is included in the candidate vehicles, and a vehicle that minimizes the customer wait time is selected to carry out the service. After finishing the service, the SOC of the battery is checked and the vehicle starts idle or charging state.

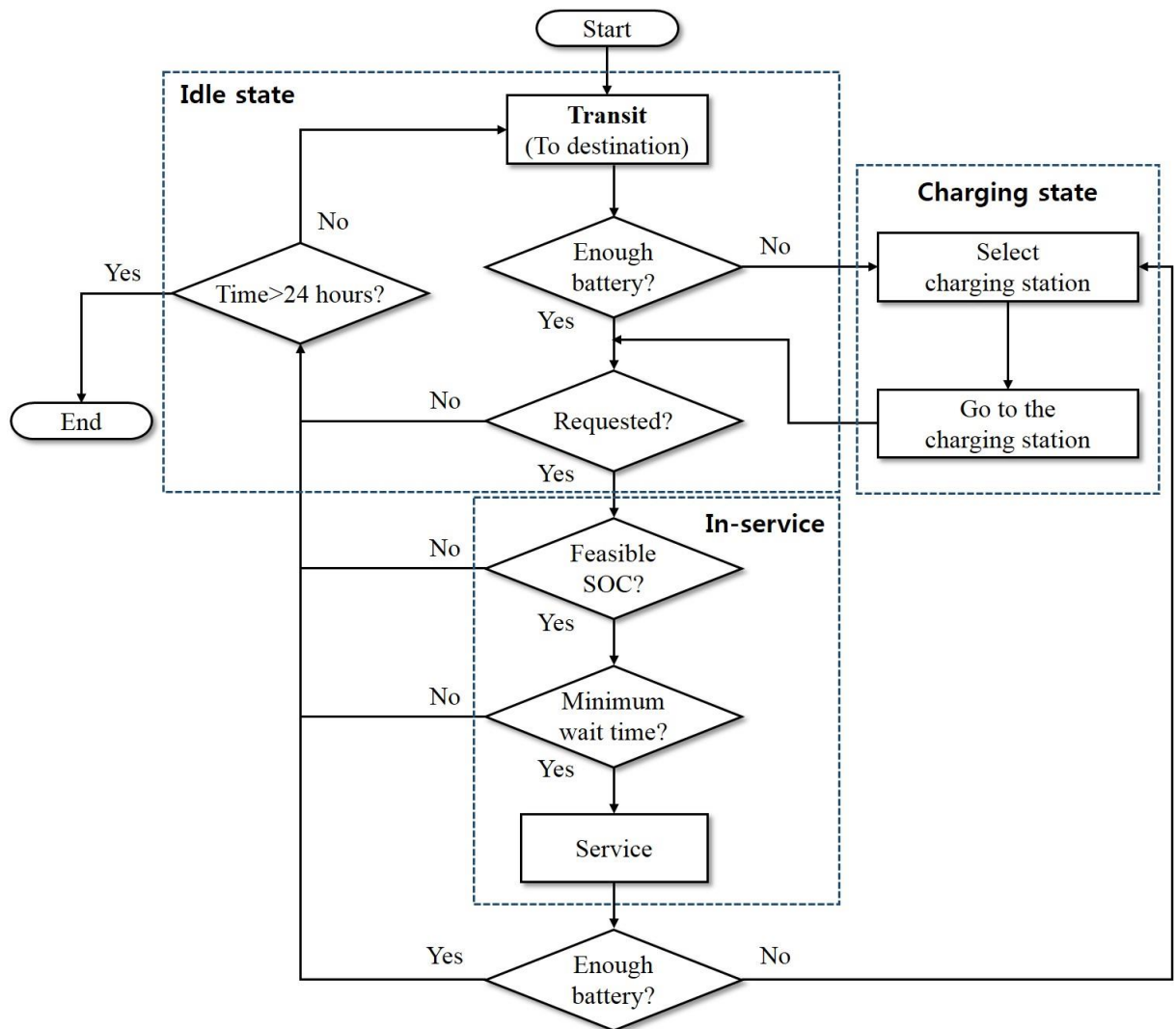

**Fig. A.1.** Flowchart of the fleet operation model.

When the SAEV moves from one place to another, it moves along the shortest time path derived from the shortest path algorithm proposed by Dijkstra [43]. The shortest time path can be obtained by considering the traffic of each road section in real time. The real-time traffic is applied to the shortest path algorithm by considering the distance of each road section differently depending on the average speed of each road section: if the average speed of the road section is lower than the average speed of Seoul roads, which is 35.8 km/h, the shortest path algorithm recognizes the longer distance of the road section with respect to the actual distance and derives the shortest path that can be regarded as the shortest time path.

Some details and information about the fleet operation model are provided next:
1. The passenger demand data, including the passenger demand time, origin, and arrival place, are generated upon the Seoul taxi data provided by the Seoul Metropolitan Government [36].
2. The central operating system is assumed to receive the passenger demand data of the customer in real time through their smartphone app.
3. If the origin or arrival place submitted by the customer is not exactly on the node, the SAEV is assumed to move to the closest node to the request point; then, the SAEV gets out of the segment (road) to reach the request point.
4. An SAEV moves in fully autonomous driving mode along the shortest time path.

## A2. SAEV design and CS models

The SAEV design and CS models provide vehicle performance and CS information to the fleet operation model, respectively. This predetermines the SAEV and CS design to explore and develop the destination selection algorithm effectively. The SAEV design model uses the EV simulation model presented by Lee et al. to determine the vehicle performance metrics, such as range, MPGe, acceleration (0−100), and top speed [35]. The EV simulation model reflects the vehicle's powertrain design, and the gear ratio is assumed to be 8.2, as in Nissan Leaf [44]. The number of battery cells in series determines the voltage of the battery, which affects acceleration and top speed, and the number of battery cells in parallel affects the range by determining the battery capacity. study

The CS model determines the capacity of the CSs by providing CS locations and number of chargers to the fleet operation model. In studySection 6.1, given that the SAEV simulation was performed on the basis of approximately 1000 customers per day, six CSs with four chargers in each station were assumed to be adequate to handle them. Government-operated CSs in Seoul were used as candidates for CS locations [45]. Among candidates, six CS locations shown in Fig. 10—node numbers 64,173, 251, 268, 424, and 459—were determined using the *p*-median model that minimizes the path distance between any node and its closest CS [46]. The capacity of the CSs determines the CS wait time, which is the time that an SAEV should wait for charging in case chargers of all CSs are in use. This affected the customer wait time in the SAEV simulations. Each charger uses a direct-current fast charging system that charges a 24-kWh battery in 30 minutes.

Before performing the SAEV system optimization in Section 5, the candidates of CS locations are pre-determined using the *p*-median model. Given that the number of CSs is prescribed, the node numbers of the CSs in Fig. 10 are determined according to Table A.1.

**Table A.1.** Candidates of CS locations.

| Number of CSs | Candidates of CS locations (node numbers) |
|---|---|
| 1 | 206 |
| 2 | 64,173 |
| 3 | 64,173,268 |
| 4 | 64,173,268,424 |
| 5 | 64,173,251,268,424 |
| 6 | 64,173,251,268,424,459 |
| 7 | 64,173,193,251,268,424,459 |
| 8 | 64,173,193,203,251,268,424,459 |
| 9 | 64,173,193,203,251,268,379,424,459 |
| 10 | 9,64,173,193,203,251,268,399,424,459 |
| 11 | 9,64,173,193,203,208,251,268,399,424,459 |
| 12 | 9,64,173,193,203,206,208,251,268,399,424,459 |
| 13 | 9,64,173,193,203,206,208,251,268,399,424,456,459 |
| 14 | 9,64,173,193,203,206,208,251,268,379,399,424,456,459 |
| 15 | 9,64,153,173,193,203,206,208,251,268,379,399,424,456,459 |

## A3. Traffic Condition

In this study, Seoul was used as a test-bed, as shown in Fig. 10. The crossing points and connections of roads are respectively set as nodes and segments to represent real roads. The test-bed represents the roads of Seoul with a total of 463 nodes and 845 segments. By representing real roads with connections of nodes and segments, real-time traffic of each road section can be reflected and the shortest path or the shortest time path can be realistically described.

To reflect the real-time traffic of Seoul, the real-time average speed data of Seoul roads provided by the Seoul Metropolitan Government were used as the average speed of each road section [47]. The average speed of road sections over time is summarized in Table A.2. The average speed of Seoul roads is 35.8 km/h. The traffic flow is smooth at dawn and becomes heavy during rush hour. An SAEV is assumed to move with real-time average speed in each road section when moving.

**Table A.2.** Average speed of road sections over time.

| Time | 0–30 km/h (%) | 30–60 km/h (%) | 60–90 km/h (%) | 90–120 km/h (%) | Average speed (km/h) |
|---|---|---|---|---|---|
| 0:00–1:00 | 36.92 | 44.85 | 15.74 | 2.49 | 41.2 |
| 1:00–2:00 | 24.38 | 56.45 | 14.20 | 4.97 | 44.1 |
| 2:00–3:00 | 16.09 | 64.50 | 14.20 | 5.21 | 46.0 |
| 3:00–4:00 | 10.76 | 68.88 | 15.27 | 5.09 | 47.4 |
| 4:00–5:00 | 11.36 | 67.93 | 15.62 | 5.09 | 47.3 |
| 5:00–6:00 | 19.53 | 60.82 | 15.86 | 3.79 | 44.8 |
| 6:00–7:00 | 30.53 | 53.61 | 15.62 | 0.24 | 40.6 |
| 7:00–8:00 | 46.86 | 41.30 | 11.72 | 0.12 | 36.3 |
| 8:00–9:00 | 60.83 | 29.35 | 9.70 | 0.12 | 33.4 |
| 9:00–10:00 | 60.95 | 29.23 | 9.70 | 0.12 | 33.3 |
| 10:00–11:00 | 61.30 | 29.35 | 9.23 | 0.12 | 32.7 |
| 11:00–12:00 | 62.01 | 28.88 | 8.87 | 0.24 | 32.7 |
| 12:00–13:00 | 60.35 | 28.05 | 11.36 | 0.24 | 33.6 |
| 13:00–14:00 | 61.18 | 27.81 | 10.77 | 0.24 | 32.8 |
| 14:00–15:00 | 64.26 | 27.10 | 8.52 | 0.12 | 31.2 |
| 15:00–16:00 | 65.91 | 26.15 | 7.81 | 0.12 | 30.4 |
| 16:00–17:00 | 66.62 | 26.51 | 6.75 | 0.12 | 29.4 |
| 17:00–18:00 | 71.01 | 23.31 | 5.68 | 0.00 | 27.4 |
| 18:00–19:00 | 74.20 | 22.49 | 3.31 | 0.00 | 26.0 |
| 19:00–20:00 | 69.47 | 26.51 | 4.02 | 0.00 | 28.3 |
| 20:00–21:00 | 61.06 | 27.93 | 11.01 | 0.00 | 32.5 |
| 21:00–22:00 | 57.04 | 29.82 | 13.02 | 0.12 | 34.4 |
| 22:00–23:00 | 54.20 | 32.31 | 13.25 | 0.24 | 35.1 |
| 23:00–24:00 | 46.86 | 37.16 | 14.91 | 1.07 | 38.1 |

## A4. Characteristics of Relocation Model

To analyze the characteristics of $f_1$ and $f_2$, function value comparison for the same $O_{ij}$—distance from current location to the destination for $f_1$ and vehicle excess at the destination for $f_2$—was performed. The results in terms of average function value for seven days are presented in Table A.3. The results show that $f_1$ gradually decreases as the distance to the destination increases, while $f_2$ rapidly decreases when vehicle excess occurs and converges to 0 when vehicle excess increases. This verifies that the passenger demand prediction model is well applied to the destination selection algorithm, avoiding assignment of idle state SAEVs in locations where the number of vehicles already assigned exceeds the expected customer demand.

Table A.3. Function value comparison for the same $O_{ij}$.

|       | $O_{ij}=1$ | $O_{ij}=2$ | $O_{ij}=3$ | $O_{ij}=4$ | $O_{ij}=5$ |
|-------|--------|--------|--------|--------|--------|
| $f_1$ | 0.8952 | 0.7536 | 0.5990 | 0.4508 | 0.3235 |
| $f_2$ | 0.1934 | 0.0123 | 0.0009 | 0.0000 | 0.0000 |

Sensitivity analysis for the probability of selection was performed to determine how $P_1$, $P_2$, and $P_3$ affect the destination selection algorithm. The exponential-Gauss function with optimal parameters was used for both $f_1$ and $f_2$. The mean wait time results depending on the combination of probability of selection are summarized in Table A.4. In the simulations, the destination selection algorithm selects a destination considering the given combination of probability of selection. When only $P_1$ is considered, the mean wait time is reduced by 59.71% (6.97 minutes) on average with respect to applying the random-motion strategy, indicating that $P_1$ provided by the passenger demand prediction model has a significant impact on the reduction of the customer wait time. Compared with the case where only $P_1$ is considered, the mean wait time is reduced by 17.33% (0.79 minutes) and 25.4% (1.17 minutes) on average in cases where $P_1$ and $P_2$ are considered, and where $P_1$ and $P_3$ are considered, respectively. These results demonstrate that $P_3$ is relatively more important than $P_2$ in the destination selection algorithm and that it is important to predict the vehicle excess at the destination accurately and assign the appropriate number of SAEVs to each destination. When all $P_1$, $P_2$, and $P_3$ are considered, the mean wait time is reduced by 33.83% (1.58 minutes) on average compared with the case where only $P_1$ is considered, showing that it is important to consider both the distance from the current location to the destination and the vehicle excess at the destination.

Table A.4. Mean wait time results depending on the combination of probability of selection (minutes).

| Combination of probability of selection | Day 1 | Day 2 | Day 3 | Day 4 | Day 5 | Day 6 | Day 7 |
|---|---|---|---|---|---|---|---|
| $P_1$ | 4.68 | 4.51 | 4.23 | 4.39 | 4.43 | 5.68 | 5.07 |
| $P_1 \times P_2$ | 4.17 | 3.77 | 3.06 | 3.43 | 3.51 | 5.20 | 4.31 |
| $P_1 \times P_3$ | 3.56 | 3.30 | 2.78 | 3.12 | 3.14 | 4.78 | 4.11 |
| $P_1 \times P_2 \times P_3$ | 3.16 | 2.95 | 2.54 | 2.76 | 2.85 | 4.16 | 3.55 |